\documentclass{article}

\usepackage[preprint]{neurips_2024}
\usepackage[utf8]{inputenc} 
\usepackage[T1]{fontenc}    
\usepackage{hyperref}       
\usepackage{url}            
\usepackage{booktabs}       
\usepackage{amsfonts}       
\usepackage{nicefrac}       
\usepackage{microtype}      
\usepackage{xcolor}         
\usepackage{graphicx}    
\usepackage{booktabs}
\usepackage{hyperref}
\usepackage{caption} 
\title{RDBE: Reasoning Distillation-Based Evaluation Enhances Automatic Essay Scoring}

\author{%
  Ali Ghiasvand Mohammadkhani  \\
  Shahid Soltani 4 High School\\
  Karaj, Iran \\
  \texttt{aghiasvandm@gmail.com} \\
}

\begin{document}

\maketitle

\begin{abstract}
  Recently,  various encoder-only and encoder-decoder pre-trained models like BERT and T5 have been applied to automatic essay scoring (AES) as small language models. However, existing studies have primarily treated this task akin to a classification problem, focusing solely on outputting scores in the target text without offering interpretations for the generated scores. Departing from the approaches, we introduce Reasoning Distillation-Based Evaluation (RDBE), which integrates interpretability to elucidate the rationale behind model scores while enhancing performance through initial reasoning. This interpretive capability is acquired during training by leveraging generated reasoning from a large language model (LLM) to distill a small language model (SLM). Our experimental results demonstrate the efficacy of RDBE across all scoring rubrics considered in the dataset. RDBE outperforms both zero-shot LLM generation and generation from a baseline fine-tuned model, establishing itself as state-of-the-art in the corresponding dataset. This highlights its practical interpretative output and enhanced performance.\footnote{\footnotesize{Implemented code released at  \url{https://github.com/AliGhiasvand86/RDBE}}}
\end{abstract}

\section{Introduction}

Automated essay scoring (AES) stands at the forefront of modern educational assessment, offering expedient evaluations of extensive essay collections. This technology plays a pivotal role in providing instantaneous feedback to students and educators alike, revolutionizing the educational landscape by facilitating timely insights into writing proficiency.

Recently, there has been a noticeable surge in holistic AES methodologies incorporating fine-grained criteria and leveraging pre-trained language models. However, adapting these models, especially advanced ones like GPT-3.5-turbo, to accommodate diverse scoring rubrics poses a significant challenge. Presently, AES predominantly relies on smaller encoder-only or encoder-decoder models for their adeptness in learning structural nuances and their cost-effectiveness in terms of training and inference, contrasting with the higher computational demands associated with large language models (LLMs). According to our final model, which is a very low-weight LLM model, it can be used for edge devices and run on not-so-powerful local machines like mobiles or personal PCs. Based on this capability, this work can be considered to have a significant social impact.

To reference these approaches, \citep{wang2022use, yang2020enhancing} have demonstrated promising results by employing BERT \citep{devlin2019bert}, while \citep{kumar2022many} introduced a multi-trait scoring approach. Additionally, ArTS proposed by \citep{do2024autoregressive}, which serves as the primary baseline for this study, utilized an autoregressive architecture with the T5 \citep{raffel2020exploring} model for scoring essays. These studies highlight diverse strategies in automated essay scoring, each leveraging specific models and methodologies to achieve notable advancements in the field.

However, to the best of our knowledge, since most previous works constructed their target text as a simple score without providing any kind of interpretation, applying reasoning distillation-based evaluation to compel the model for reasoning and providing interpretation through the AES process was not discovered, and we have entered into this intuition with two objectives. The first is to understand the logic behind the model's decision to give that score, and the second one is to improve scoring performance based on proper reasoning and interpretation. 
Towards this end, we propose a reasoning distillation-based evaluation framework for AES, making several key contributions: \\ \\
1.	We introduce RDBE, a simple yet effective AES framework that integrates learning how to reason and interpret essay evaluation during training and automates the same process during inference. Additionally, RDBE’s principles can be adapted for other tasks involving the evaluation of long-form generated text by customizing tasks and scoring rubrics. \\ \\
2.	By leveraging the DREsS\(_{New}\) dataset \citep{yoo2024dress} as a foundation and creating reasoning and interpretation for the score of each data point across various scoring rubrics using the Llama-3-70B model, we propose a novel and effective dataset for fine-tuning models in reasoning distillation-based essay evaluation. \\ \\
3.	Experimental results demonstrate that our reasoning and interpretation not only enhance interpretative output significantly but also achieve state-of-the-art performance on the DREsS\(_{New}\) dataset across all scoring rubrics, surpassing competitive baselines.

\section{Related Works}

\subsection{Automatic Essay Scoring}
Automated Essay Scoring (AES) is a crucial field connecting natural language processing (NLP) and education. Early AES approaches were primarily based on regression or classification machine learning models, using textual features from essays \citep{salim2019automated}. Recent advancements have seen the integration of deep learning techniques like convolutional neural networks (CNNs) \citep{dong2016automatic} and pre-trained language models, enhancing scoring accuracy. Recent advancements in automated essay scoring (AES) often leverage BERT and T5, significantly enhancing AES capabilities.

\subsection{Reasoning Distillation}
Recent studies have concentrated on extracting the logical reasoning abilities of large language models (LLMs) into smaller-scale models \citep{ho2023large, hsieh2023distilling}. These approaches involve extracting and utilizing rationales behind LLM predictions to enhance the training of smaller models, thereby boosting their overall performance.

\section{Methodology}
\begin{figure}
  \centering
  \begin{center}
  %
  \includegraphics[width=\dimexpr\textwidth-2\fboxrule-2\fboxsep]{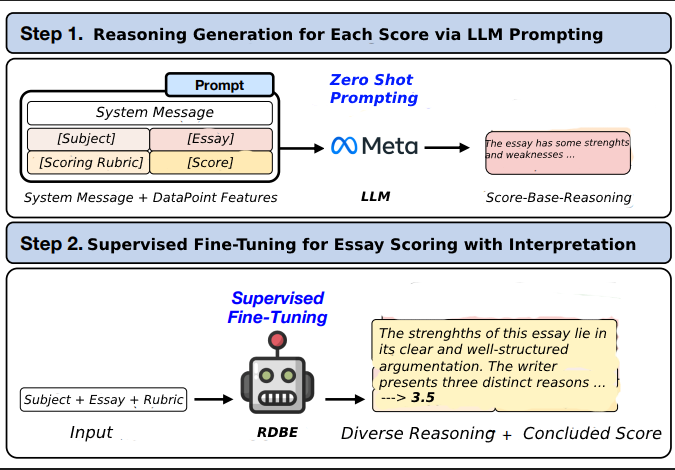}

  %
  \caption{Automated Essay Scoring Framework:
Step 1 involves using zero-shot prompting via a language model (LLM) to generate reasoning for each input prompt containing the subject, essay, scoring rubric, and score.
Step 2 entails supervised fine-tuning to produce diverse reasoning and a concluded score, refining the essay evaluation process.}
  \label{framework}
  \end{center}
\end{figure}

Our method involves utilizing two steps to form the whole framework. In the first step, we consider the DREsS\(_{New}\) dataset and also consider the same conceptual descriptions used in each of the three scoring rubrics for content, organization, and language mentioned in the DREsS paper. The main definition of this problem is to score a received essay, given its subject and the specified scoring rubric, based on the score from the list [1, 1.5, 2, 2.5, 3, 3.5, 4, 4.5, 5]. An illustration of our two-step interpretable framework is depicted in Figure~\ref{framework}.

\subsection{Synthetic Data Generation}
For each data point, as three scoring rubrics are considered, we prompt the Llama-3-70B model using the Groq API\footnote{\footnotesize{\url{https://groq.com/}}} to generate a score for each rubric. We then instruct the model to reason and interpret why the essay received that specific score for each scoring rubric. The generated reasoning from the LLM for each scoring rubric for each data point is then considered as a new output for further steps. As shown in the first step of the Figure~\ref{framework}, we use the system message below. For the user message, we prompt the LLM with the corresponding subject, essay, scoring rubric, and score exactly in the order written in this sentence, each coming after its respective tag ([Subject], [Essay], [Scoring Rubric], or [Score]), so the model can generate reasoning for the achieved score. The system message is:

"You are a helpful, accurate, and fair virtual assistant with expertise in English composition, specializing in evaluating and accurately scoring essays written by students. A predefined [Subject] is provided, followed by the corresponding [Essay]. A [Scoring Rubric] is then explained, and the [Score] that the essay receives is based on the mentioned rubric without considering any other rubrics, even those logical or required, represented by a number from the list [1.0, 1.5, 2.0, 2.5, 3.0, 3.5, 4.0, 4.5, 5.0]. As an expert in accurate essay evaluation, your role is to provide insights into strengths and weaknesses related to the score by analyzing only both the quality of presence (if present) or absence of the aspects directly mentioned in the rubric's explanation within the essay based on the received score and interpret them in a short text. For scores higher than 2.5, you should focus more on identifying strengths and give less attention to stating weaknesses. Conversely, for scores lower than 2.5, prioritize identifying weaknesses and give less attention to stating strengths."

\subsection{Fine-tuning Backbone Model}
In this step, we fine-tune our backbone encoder-decoder base transformer model using the data obtained from the previous step. The input is structured as follows: first, the scoring rubric is explained after a [Scoring Rubric] tag, then the essay subject is written after a [Subject] tag, and finally, the body of the essay is written after a [Essay] tag. The output consists of the corresponding reasoning and interpretation generated by the LLM, followed by “ ---> ” and the score for that rubric. This setup (second step of the Figure~\ref{framework}) is designed to train the model to generate the score after providing its reasoning and interpretation.

\section{Experiments and Results}
In this section, we provide a comprehensive analysis of the utilized dataset, evaluation metrics, main achieved results, and evaluated baselines.

\subsection{Datasets and Settings}
For the main experiment, we utilized the DREsS\(_{New}\) dataset, which consists of 1980 data points after removing invalid and NaN data (due to ongoing development issues with the dataset). Each data point includes the subject of the essay, the body of the essay, and scores for content, organization, language, and the sum of these three scores. The dataset was split into 60\% for training, 20\% for development, and 20\% for testing using a random seed of 22.

During inference with the Llama-3-70B model, a temperature of 0 was set to ensure reproducibility. For both our approach and the baseline \citep{do2024autoregressive}, we employed the LongT5-Base model \citep{guo2021longt5} from the HuggingFace transformers library\footnote{\footnotesize{\url{https://github.com/huggingface/transformers}}} as the backbone model for fine-tuning. This model utilizes transient-global attention and has approximately 220 million parameters, allowing it to handle longer input sizes compared to standard T5 models, as LongT5 models are pre-trained on inputs with more tokens. For fine-tuning, we used the AdamW optimizer and the cross-entropy loss function. The model was trained for 15 epochs with a batch size of 8.

\begin{table}[t]
    \centering
    \begin{tabular}{lc|ccc|c}
        \toprule
        Model & Fine-tuning Data & Content & Organization & Language & Total \\
        \midrule
        Llama-3-70B & N/A & 0.167 & 0.108 & 0.090 & 0.119 \\
        \midrule
        ArTS \citep{do2024autoregressive} & DREsS\(_{New}\) & 0.516 & 0.508 & 0.535 & 0.559 \\
        RDBE & DREsS\(_{New}\) & \textbf{0.606} & \textbf{0.629} & \textbf{0.638} & \textbf{0.730} \\
        \bottomrule
    \end{tabular}
    \caption{QWK scores of our framework and two baselines in the three specified scoring rubrics, plus the total score: the first is Llama-3-70B in zero-shot, and the second is ArTS with its proposed strategy implemented for customized evaluation on the DREsS\(_{New}\) dataset.}
    \label{tab:model_comparison}
\end{table}

\subsection{Main Results}
For evaluation, we used the quadratic weighted kappa (QWK) \citep{cohen1968weighted}, which is the official metric for the dataset.
As mentioned in Table ~\ref{tab:model_comparison}, we assessed our framework on the test split of the DREsS\(_{New}\) dataset separately for each of the three scoring rubrics and also for the total overall score. Additionally, we evaluated two strong baselines: the Llama-3-70B model in a zero-shot manner based on the prompt template "C" mentioned in the DREsS paper, and ArTS, as a a autoregressive scoring baseline on this dataset. However, our framework consistently achieved better QWK scores across all evaluations, outperforming Llama-3-70B despite having only 0.03\% of its parameters and also performing better than ArTS, a novel and effective autoregressive strategy. Consequently, our framework reaches a state-of-the-art performance on this dataset.

\section{Conclusion and Future Work}
This study introduces RBDE, a novel and straightforward framework for interpretable automatic essay scoring. It provides a new dataset based on reasoning generated from a large language model, followed by training a model to distill reasoning and improve interpretability in essay evaluation. Based on the results, our approach of enhancing evaluation capability for small language models through reasoning distillation can be applied to other evaluation tasks, suggesting promising directions for future research in this field. 

However, the main limitation of our work was the quality of our synthesized data. Due to our limited budget, we were unable to utilize more accurate models like GPT-4 for data synthesis, which could have generated higher-quality data and improved the performance of small language models in interpretation and scoring.

\begin{ack}
I conducted all the research, coding, and development of ideas and contributions for this project. Ali Royat, a PhD student, supervised me, providing mentoring and facilitating discussions. I am thankful to him. Furthermore, there are no financial activities supporting the submitted work.
\end{ack}

Also, LLMs were never used for our research process, including developing ideas and contributions. They were only used to refine the language and grammar of our paper and not to write the paper from scratch.

\medskip

\bibliography{bibliography}




\end{document}